\documentclass[final]{cvpr}

\usepackage{times}
\usepackage{epsfig}
\usepackage{graphicx}
\usepackage{amsmath}
\usepackage{amssymb}

\usepackage[pagebackref=true,breaklinks=true,colorlinks,bookmarks=false]{hyperref}

\usepackage{multirow}
\usepackage{booktabs}

\def\cvprPaperID{2205} 
\def\confYear{CVPR 2021}

\begin{document}
\title{AutoAssign: Differentiable Label Assignment for Dense Object Detection}

\author{Benjin Zhu, Jianfeng Wang, Zhengkai Jiang, Fuhang Zong, Songtao Liu, Zeming Li, Jian Sun \\
MEGVII Technology\\
{\tt\small \{zhubenjin, wangjianfeng, jiangzhengkai, zongfuhang, liusongtao, lizeming, sunjian\}@megvii.com}
}

\maketitle

\begin{abstract}
Determining positive/negative samples for object detection is known as label assignment. Here we present an anchor-free detector named AutoAssign. It requires little human knowledge and achieves appearance-aware through a fully differentiable weighting mechanism. During training, to both satisfy the prior distribution of data and adapt to category characteristics, we present Center Weighting to adjust the category-specific prior distributions. To adapt to object appearances, Confidence Weighting is proposed to adjust the specific assign strategy of each instance. The two weighting modules are then combined to generate positive and negative weights to adjust each location's confidence. Extensive experiments on the MS COCO show that our method steadily surpasses other best sampling strategies by large margins with various backbones. Moreover, our best model achieves 52.1\% AP, outperforming all existing one-stage detectors. Besides, experiments on other datasets, \emph{e.g.}, PASCAL VOC, Objects365, and WiderFace, demonstrate the broad applicability of AutoAssign.

\end{abstract}

\section{Introduction}

Current state-of-the-art CNN based object detectors perform a common paradigm of dense prediction. Both two-stage (the RPN~\cite{ren2015faster} part) and one-stage detectors~\cite{lin2017focal,tian2019fcos,zhang2019freeanchor,zhang2019bridging} predict objects with various scales, aspect ratios, and classes over every CNN feature locations in a regular, dense sampling manner. This dense detection task raises an essential issue of sampling positives and negatives in the spatial locations, which we call \textit{label assignment}. Moreover, as the modern CNN-based detectors commonly adopt multi-scale features (\emph{e.g.}, FPN~\cite{lin2017feature}) to alleviate scale variance, label assignment requires not only selecting locations among spatial feature maps (\textit{spatial assignment}) but also choosing the level of features with appropriate scale (\textit{scale assignment}).  

\begin{figure}[htbp]
\begin{center}
\begin{tabular}{ccc}
\includegraphics[width=\linewidth]{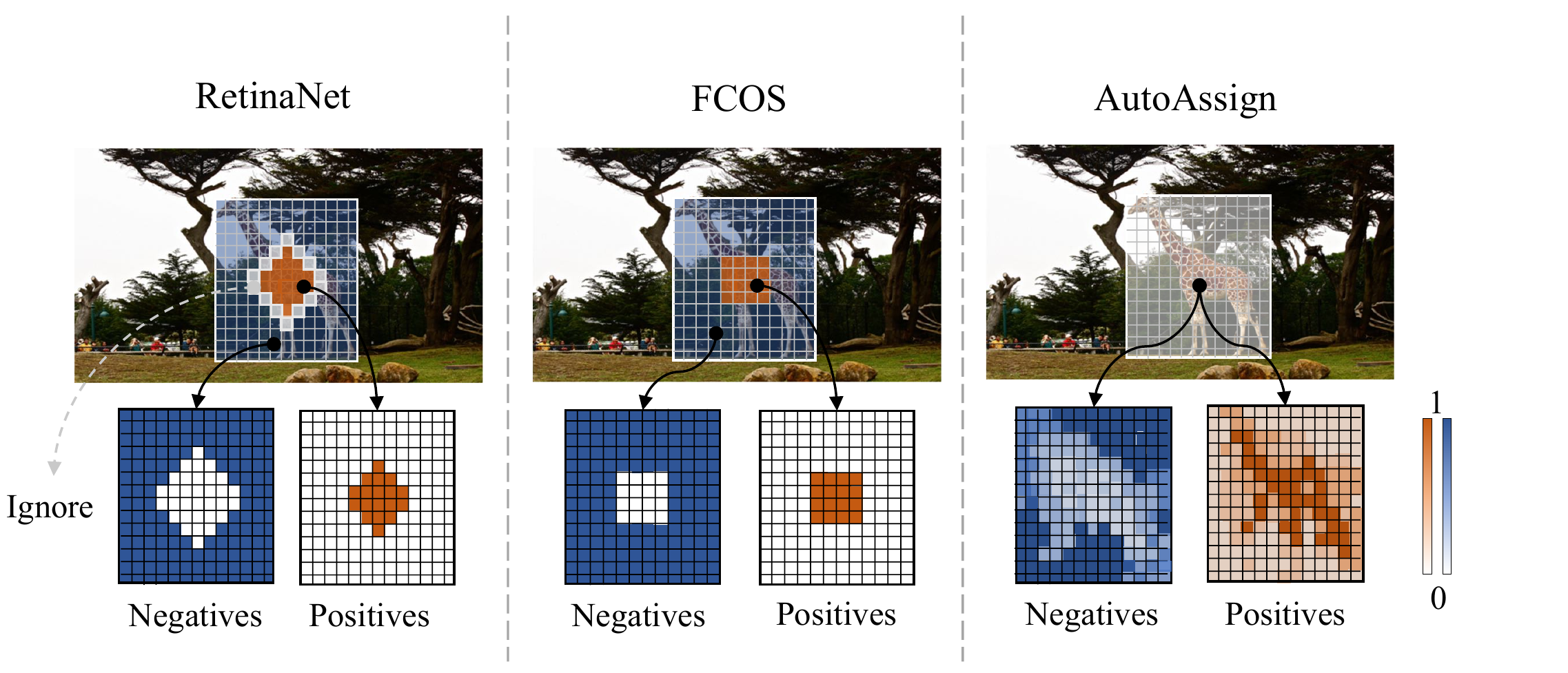}
\end{tabular}
\end{center}
\caption{Illustration of different label assignment strategies. Compared to fixed label assignment strategies like RetinaNet and FCOS, AutoAssign do not rely on preset samples and can adapt to object appearance automatically. For better visualization, we stack locations across multiple scales to show the final results.}
\label{fig:fig1_illustration}
\end{figure}

As shown in Fig.~\ref{fig:fig1_illustration}, existing detectors mainly sample the positive and negative locations by human prior knowledge: (1) Anchor-based detectors like RetinaNet~\cite{lin2017focal} preset several anchors of multiple scales and aspect ratios on each location and resort to the Intersection over Union (IoU) for sampling positives and negatives among spatial and scale-level feature maps. (2) Anchor-free detectors like FCOS~\cite{tian2019fcos} sample a fixed fraction of center area as positive spatial locations for each object, and select certain stages of FPN~\cite{lin2017feature} by the pre-defined scale constraints. These detectors follow the \textit{center prior} (objects are more likely to located around the center of their bounding box) in data distributions to design their assignment strategies, which are proved to be effective on benchmarks like Pascal VOC~\cite{pascal-voc-2007,pascal-voc-2012} and MS COCO~\cite{lin2014microsoft}. However, appearances of objects vary a lot across \textit{categories} and \textit{scenarios}. The above fixed center sampling strategy may pick locations outside objects (\emph{e.g.}, bananas, umbrellas) as positives, thus cannot cover the diverse distributions of categories. 

To deal with the diverse data distributions, a few recent works introduce some partially dynamic strategies in label assignment. GuidedAnchoring~\cite{wang2019region} and MetaAnchor~\cite{yang2018metaanchor} dynamically change the prior of anchor shapes before sampling, while other methods adaptively modify the sampling strategy for each object in the spatial dimension~\cite{zhang2019freeanchor,zhang2019bridging,noisyanchor} or the scale dimension~\cite{zhu2019feature}. The success of these partially dynamic methods demonstrates great potential in making label assignment more adaptive. However, these strategies can only free part of the label assignment to be data-driven. The other parts stay constrained by human designs, preventing label assignment to be further optimized.


Intuitively, sampling locations on objects is better than background because they are prone to generate higher quality proposals. Motivated by this, we present AutoAssign, which makes label assignment fully data-dependent and appearance-aware. By dropping the many human knowledge (\emph{e.g.}, anchors, IoU thresholds, and top-$k$) and proposing a unified weighting mechanism across spatial and scales, we reach a  fully differentiable strategy.
 
We adopt a similar paradigm of anchor-free detectors like FCOS~\cite{tian2019fcos} to predict one object proposal at each location directly. Given an object, 
we initially treat all the locations across FPN scales inside its bounding box as \textit{both} positive and negative candidates for further optimization. To adapt to the data distribution of different categories, we propose a category-wise Center Weighting module to learn each category's distribution. To get adapted to each instance's appearance and scale, we propose a Confidence Weighting module to modify the positive and negative confidences of the locations in both spatial and scale dimensions. The two weighting modules are combined to generate positive and negative weight maps for all locations inside an object. According to Fig.~\ref{fig:fig1_illustration}, the assignment results can dynamically adapt to object appearances. 
The entire process of weighting is differentiable and can be conveniently optimized by back-propagation during training. All of the weighting modules are only used during loss calculation; thus, AutoAssign is inference cost-free. Moreover, the proposed method only requires the \textit{center prior} knowledge, saving a lot of effort in hyper-parameters tuning, thus can accommodate other data distributions conveniently without any modification.

In summary, the contributions of this study are three-fold as follows:

\begin{enumerate}
	\item An appearance-aware and fully differentiable weighting mechanism for label assignment is proposed. It enables spatial and scale assignment to be optimized in a unified manner.
	\item Two weighting modules (\emph{i.e.}, Center Weighting and Confidence Weighting) are proposed to adjust the category-specific prior distribution and the instance-specific sampling strategy in both spatial and scale dimensions. 
	\item AutoAssign achieves state-of-the-art performance on the challenging MS COCO~\cite{lin2014microsoft} dataset. Competitive results on datasets from different distributions, such as PASCAL VOC~\cite{pascal-voc-2007,pascal-voc-2012}, Object365~\cite{shao2019objects365} and WiderFace~\cite{yang2016wider} demonstrate the effectiveness and broad applicability of AutoAssign.
\end{enumerate}

\section{Related Work}
\label{sec:related_work}

\paragraph{Fixed Label assignment}

Classical object detectors sample positives and negatives with pre-defined strategies. The RPN in Faster R-CNN~\cite{ren2015faster} preset anchors of different scales and aspect ratios at each location. Given an instance, assignments in both scale and spatial dimensions are guided by the anchor matching IoU. This anchor-based strategy quickly dominates modern detectors and extends to multi-scale outputs (\emph{e.g.}, YOLO~\cite{redmon2017yolo9000,redmon2018yolov3}, SSD~\cite{liu2016ssd}, and RetinaNet~\cite{lin2017focal}). Recently, attention has been geared toward anchor-free detectors. FCOS~\cite{tian2019fcos} and its precursors~\cite{huang2015densebox,yu2016unitbox,redmon2016you} drop the prior anchor settings and directly assign the spatial positions around bounding box center of each object as positives. In scale dimension, they pre-define scale ranges of different FPN~\cite{lin2017feature} stages to assign instances of different sizes. Both the anchor-based and anchor-free strategies follow the  \textit{center prior} inherent in data distributions
. However, all of these methods only depend on human knowledge to solve spatial and scale assignment separately and cannot adapt to instance appearances.


\paragraph{Dynamic Label assignment}
Recent detectors propose adaptive mechanisms to improve label assignment. GuidedAnchoring~\cite{wang2019region} leverages semantic features to guide the anchor settings and dynamically change the shape of anchors to fit various distributions of objects. MetaAnchor~\cite{yang2018metaanchor} randomly samples anchors of any shapes during training to cover different kinds of object boxes. Besides the modification of anchor prior, some works directly change the sampling for each object. FSAF~\cite{zhu2019feature} dynamically assigns each instance to the most suitable FPN feature level with minimal training loss. SAPD~\cite{zhu2019soft} re-weights the positive anchors and applies an extra meta-net to select the proper FPN stages. FreeAnchor~\cite{zhang2019freeanchor} constructs a bag of top-$k$ anchor candidates based on IoU for every object and uses a Mean-Max function to weight among selected anchors, and NoisyAnchor~\cite{noisyanchor} designs another weighting function to eliminate noisy anchors. ATSS~\cite{zhang2019bridging} proposes an adaptive training sample selection mechanism by the dynamic IoU threshold according to the statistical characteristics of instances. Concurrent work PAA~\cite{kim2020probabilistic} adaptively separates anchors into positive and negative samples in a probabilistic manner. However, they still rely on hand-crafted anchors, thresholds, or other human knowledge for guiding the assignment, which could prevent label assignment from being further optimized.


\section{Methodology}
 
 \begin{table*}
	\begin{center}
	\begin{tabular}{l|c|c|c|c}
	\toprule
	\multirow{2}{*}{\textbf{Method}} & \multirow{2}{*}{\textbf{Prior}} & \multicolumn{2}{c|}{\textbf{Instance}} & \multirow{2}{*}{\textbf{\textit{AP}}} \\ \cline{3-4}
	 &  & \textbf{scale} & \textbf{spatial} & \\
	\midrule
	RetinaNet~\cite{lin2017focal} & anchor & size \& IoU & IoU & 36.3 \\
	FreeAnchor~\cite{zhang2019freeanchor} & anchor & size \& IoU & top-$k$ weighting, IoU & 38.7 \\
	ATSS~\cite{zhang2019bridging} & anchor & size \& IoU & top-$k$, dynamic IoU & 39.3 \\
	GuidedAnchoring~\cite{wang2019region} & dynamic anchor & size \& IoU & IoU & 37.1 \\
	FCOS*~\cite{tian2019fcos} & center & range & radius & 38.7 \\
	FSAF~\cite{zhu2019feature} & anchor \& center & loss & IoU \& radius & 37.2 \\
	\midrule
	\textbf{AutoAssign (Ours}) & \textbf{Center Weighting} & \multicolumn{2}{c|}{\textbf{Confidence Weighting}} &  \textbf{40.5} \\
	\bottomrule
	\end{tabular}
	\end{center}
	\caption{Comparison of label assignment between different typical detectors. Results in terms of $AP$ (\%) are reported on the MS COCO 2017 \textit{val} set,  using ResNet-50~\cite{he2016deep} as backbone. * denotes improved versions.}
	\label{tab:comparison_of_label_assignment}
\end{table*}

Before starting, we need to ask: \textit{which part of  label assignment is essential?} To answer the question, we present existing label assignment strategies from a more holistic perspective in Table~\ref{tab:comparison_of_label_assignment}. We organize the components of some representative methods as \textit{prior}-related and \textit{instance}-related. Clearly, apart from the heuristic-based methods like RetinaNet~\cite{lin2017focal} and FCOS~\cite{tian2019fcos}, all the existing dynamic strategies benefit from its dynamic parts. But they only make partial components of label assignment data-driven, and the other components still rely on hand-crafted rules. We can conclude that: (1) All of the existing detectors obey the \textit{center prior}. Sampling locations near box centers is effective.
(2) Both spatial and scale assignments need to be tackled. But existing methods all solve the scale and spatial assignments using two different strategies.

Motivated by these observations, our aim becomes making both the \textit{prior}-related and \textit{instance}-related components adapt to the category or instance characteristics. In this section, we will first give an overall picture of AutoAssign, then demonstrate how the \textit{prior}- and \textit{instance}-level tasks are solved.

\subsection{Overview}

\begin{figure*}[htbp]
	\centering
	\includegraphics[width=0.9\linewidth]{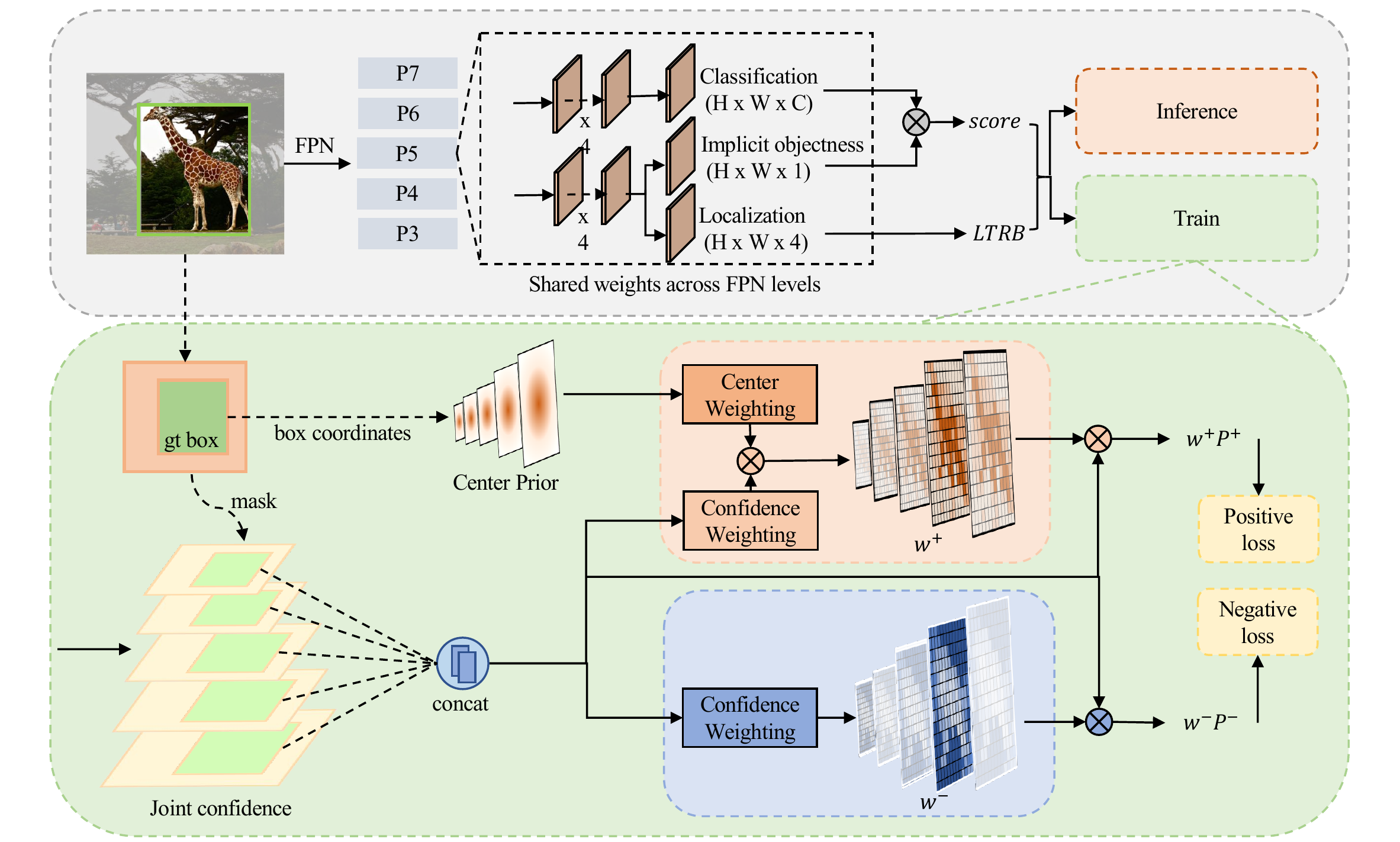}
	
\caption{Illustration of AutoAssign. The upper block shows network architecture. The product of classification and \textit{ImpObj} is used as final classification confidence. LTRB means the localization offsets are in \textit{left-top-right-bottom} format. The bottom block presents the label assignment strategy. Given an object, its box coordinates are used for calculating the initial \textit{center prior} and generating foreground masks to select inbox locations. The indexed locations will be flattened and concatenated together. For positive candidates, both Confidence Weighting and Center Weighting are used. For negative candidates, only Confidence Weighting is applied. As a result, positive and negative weight maps are generated. In this process, both spatial and scale assignments are finished jointly. 
}
\label{fig:fig2_pipeline}
\end{figure*}
As shown in Fig.~\ref{fig:fig2_pipeline}, the upper gray box shows network architecture.
We first follow the anchor-free manner like FCOS \cite{tian2019fcos} to drop the pre-designed anchors and directly predict objects on each feature location. The network has three outputs: classification score, \textit{Implicit-Objectness (ImpObj)} score (which will be described later), and localization offsets. During training (the bottom green box), we first convert all the network predictions into a joint confidence indicator. On top of this, we propose a weighting mechanism, which consists of a Center Weighting module and a Confidence Weighting module. The Center Weighting module is designed to both satisfy the inherent \textit{center prior} property in data and adapt to each category's specific shape pattern. It starts from the standard \textit{center prior} and then learns the distribution of each category from data. The Confidence Weighting module is for assigning the most appropriate locations of each instance based on its appearance and scale adaptively. For each location $i$ in a ground-truth (gt) box, The two modules are combined together to generate positive and negative weights. Finally, positive and negative classification loss will be calculated, and label assignment will be optimized jointly with the network. 

From the label assignment perspective, given an object, AutoAssign can automatically find both its appropriate scales across FPN levels and spatial locations based on the network outputs. As a result, the task of label assignment is solved properly in a unified, appearance-aware, and differentiable manner.

\subsection{Prior-level: Center Weighting}

The prior distribution is a fundamental element for label assignment, especially in the early stage of training. In general, the distribution of objects is subject to the \textit{center prior}. However, the objects from different categories, \emph{e.g.}, giraffe, and human, may have distinct distributions. Keeping sampling center positions cannot capture the diverse distributions of different categories. Preferably, adaptive center distributions for different categories are more desired.

Starting from the \textit{center prior}, we introduce a category-wise Gaussian-shape weighting function $ G $ with learnable parameters. This Center Weighting module guarantees that locations closer to bounding box center have higher location weights than locations far from box center. Moreover, it can automatically adjust its shape according to data distributions of different categories. Here we define $G$ as:
 \begin{align}\label{eq:gau}
 	G(\vec{d}\mid \vec{\mu}, \vec{\sigma}) = e^{\frac{-(\vec{d}-\vec{\mu})^2}{2\vec{\sigma}^2}},
 \end{align}
where $\vec{d}$ denotes the offsets of a certain position inside an object to its box center along $x$- and $y$-axis, which means it can be negative. $\vec{\mu}$ and $\vec{\sigma}$ are learnable parameters of shape $(K, 2)$. $K$ is the number of categories of a dataset. Each category has two parameters along spatial dimension. As $G$ contributes to the training loss, the parameters can be optimized by back-propagation. At the beginning, $\vec{\mu}$ is initialized to 0 and $\vec{\sigma}$ to 1. Intuitively, $\vec{\mu}$ controls center offset of each category from the box center. And $\vec{\sigma}$ measures each location's importance based on category characteristics. As shown in Fig.~\ref{fig:fig2_pipeline}, the bounding box will generate a location weight map as demonstrated in ``Center Prior''.

Given an object, we calculate the location weights using $G$ on every FPN stage individually, then stack the weighting results together for later usage. Furthermore, to mitigate the interference caused by the different scales of FPN, we normalize the distance $\vec{d}$ by its downscale ratio.


\subsection{Instance-level: Confidence Weighting}
\label{sec:conf_weight}


As mentioned above, all locations inside a bounding box across FPN stages will be considered as both positive and negative sample candidates at the beginning. This operation will significantly increase the background locations in positive candidates and vice versa.  This is quite different from all existing label assignment strategies, which only sample a subset of locations as positives before loss calculation.  On the other hand, given a location inside a bounding box, to obtain a reasonable weight, all aspects, including classification and regression, need to be taken into account. 

Motivated by these aspects, in Confidence Weighting, we propose a joint confidence indicator of both classification and localization to guide the weighting strategy in both spatial and scale dimensions. 

\paragraph{Classification confidence.} 


Generally speaking, selected positive samples of typical detectors imply that these locations have high confidence of containing instances. However, in our setting, the initial positives set tends to contain a considerable part of background locations, as an object can hardly fill its bounding box completely.
Consequently, if a location is, in fact, background, all class predictions in the location should be unreasonable. So taking too many inferior background locations as positives will damage detection performance, which is also the case for the negatives set.
To suppress noisy candidates (\emph{i.e.}, backgrounds in positives set, foregrounds in negatives set) from the inferior locations, we introduce a novel \textit{Implicit-Objectness (ImpObj)} branch, which is shown in Fig.~\ref{fig:fig2_pipeline}. The form of \textit{ImpObj} is just like the \textit{center-ness} in FCOS, but here we meet another issue of lacking explicit supervisions. Considering the aim that we need to find and emphasize proper positives and filter out noise candidates dynamically, we optimize the \textit{ImpObj} together with the classification branch. Specifically, we use the product of \textit{ImpObj} and classification score as our rectified classification confidence. \textit{ImpObj} thus shares supervision with the classification branch and does not require explicit labels. 

\paragraph{Joint confidence indicator.}

For generating unbiased estimation of each location towards positives/negatives, we should include the localization confidence besides classification. The typical outputs of localization are box offsets, which are hard to measure the regression confidence directly. Considering the fact that Binary Cross-Entropy (BCE) loss is commonly adopted for classification task, we thus convert the localization loss $\mathcal{L}_i^{loc}$ into likelihood:
\begin{align}
    \label{eq:loc_conf}
	\mathcal{P}_i(loc) &  = e^{-\lambda \mathcal{L}_i^{loc}},
\end{align}
for being combined with classification confidence conviently, in which $\lambda$ is a hyper-parameter to balance between classification and localization. GIoU loss~\cite{rezatofighi2019generalized} is used as $\mathcal{L}_i^{loc}$. Then we combine classification and regression likelihood together to get the joint confidence $\mathcal{P}_i$. 
For the positive candidates, we define positive confidence
    $\mathcal{P}_i^{+} = \mathcal{P}_i(cls) \cdot  \mathcal{P}_i(loc)$,
where classification confidence $\mathcal{P}_i(cls)$ is the product of classification score and \textit{ImpObj} score.
For a location candidate in negatives set, considering the fact that only classification task will be performed on negative locations, thus the negative confidence $\mathcal{P}_i^{-}  = \mathcal{P}_i(cls)$,
which is the same as locations outside bounding boxes. Therefore, all background locations can be tackled uniformly.


%

\paragraph{Positive weights.}

If a location has higher confidence towards positive samples, we prefer to take it as a foreground. Based on the joint confidence representation $\mathcal{P}_i^+$, we thus propose our confidence weighting function $C(\mathcal{P}_i^+)$ in an exponential form to emphasize the locations with high confidence containing objects as:
\begin{align}
    \label{eq:conf_weight_func}
    C(\mathcal{P}_i^+) = e^{ \mathcal{P}_i^+ / \tau },
\end{align}
where $\tau$ is a hyper-parameter to control the contributions of high and low confidence locations towards positive losses. Intuitively, given an object $i$, for all locations inside its bounding box, we should focus on the proper locations with more accurate predictions. However, at the start of the training process, the network parameters are randomly initialized, making its predicted confidences unreasonable. Thus guiding information from prior is also critical. For location $i \in S_n$, where $S_n$ denotes all locations inside the bounding box at all the scale levels of object $n$, we combine the category-specific prior $G(\vec{d}_i)$ from center weighting module and the confidence weighting module $C(\mathcal{P}_i^+)$ together to generate the positive weights $w_i^+$ as:
\begin{align}
\label{eq:positive}
 	w_i^+ & = \frac{C(\mathcal{P}_i^+) G(\vec{d}_i)}{\sum_{j \in S_n} C(\mathcal{P}_j^+) G(\vec{d}_j)},
\end{align}
here for an object $n$, each $w_i^+$ is normalized by sum of location candidates in $S_n$ for the purpose of being used as valid weights.

\paragraph{Negative weights.}
As discussed above, a bounding box usually contains an amount of real-background locations, and we also need weighted negative losses to suppress these locations and eliminate false positives. Moreover, as the locations inside the boxes always tend to predict high confidence of positives, we prefer the localization confidence to generate the unbiased indicator of false positives. Paradoxically, the negative confidence $\mathcal{P}^-$ has no gradient for the regression task, which means the localization confidence $\mathcal{P}_i(loc)$ should not be optimized by negative loss. Hence we use IoUs between each position's predicted proposal and all objects to generate our negative weights $w_i^-$ as: 
\begin{align}
\label{eq:negative}
	w_i^- & = 1 - f(\text{iou}_i),
\end{align}
in which $f(\text{iou}_i) = 1 / (1 - \text{iou}_i)$, $\text{iou}_i$ denotes max IoU between proposal of location $i \in S_n$ and all ground truth boxes. To be used as valid weights, we normalize $f(\text{iou}_i)$ into range $[0, 1]$ by its value range. This transformation sharpens the weight distributions and ensure that the location with highest IoU receives zero negative loss. For all locations outside bounding boxes, $w_i^-$ is set to 1 because they are backgrounds for sure.

\subsection{Loss function}
By generating positive and negative weight maps, we achieve the purpose of dynamically assigning more appropriate spatial locations and automatically selecting the proper FPN stages for each instance. As the weight maps contribute to the training loss, AutoAssign tackles the label assignment in a differentiable manner. The final loss function $\mathcal{L}$ of AutoAssign is defined as follows:
\begin{equation}
\label{eq:loss_func}
	\mathcal{L}\! = \! -\sum_{n=1}^{N} \log (\sum_{i \in S_n}  w_i^+ \mathcal{P}_i^+ ) - \sum_{k \in S} \log(1 - w_k^- \mathcal{P}_k^-), 
\end{equation}
$S$ denotes all the locations at all the scales on the output feature maps. To ensure at least one location matches object $n$, we use the weighted sum of all positive weights to get the final positive confidence. Thus for a location inside bounding boxes, both positive and negative loss will be calculated with different weights. The positive loss and negative loss are calculated independently. Thus the magnitude of positive and negative weights requires no extra operation.
To handle the severe imbalance problem in negative samples, the Focal Loss~\cite{lin2017focal} is applied to the negative loss in Eq.~\ref{eq:loss_func}.


\section{Experiments}
\label{sec:exp}

\begin{figure*}[htbp]
	\centering
	\includegraphics[width=0.9\linewidth]{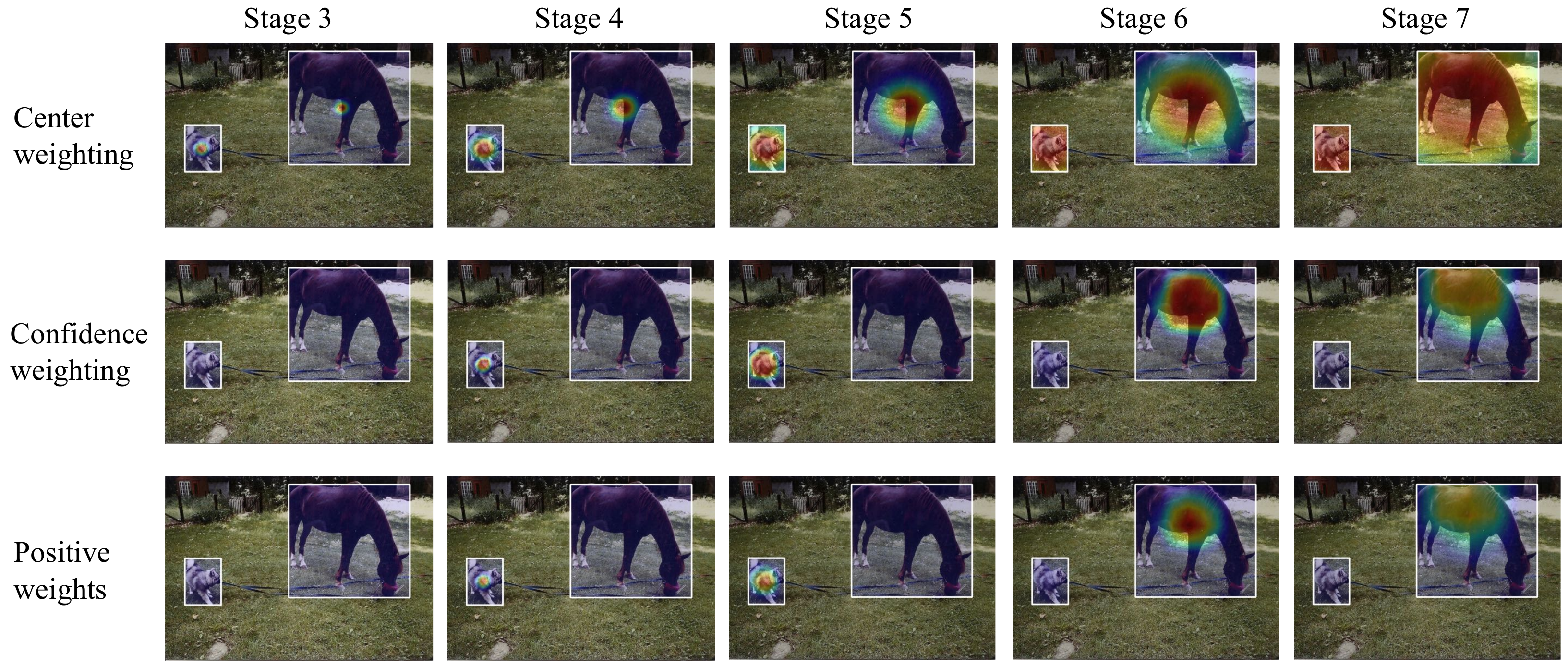}
\caption{Visualization of center weighting, confidence weighting, and positive weights. From the 3rd row, objects of different shapes and sizes are assigned to its appropriate spatial locations and suitable scale stages automatically.}
\label{fig:da_weighting}
\end{figure*}

Experiments are mainly evaluated on the MS COCO 2017~\cite{lin2014microsoft} benchmark, which contains around 118k images in the \textit{train} set, 5k in the \textit{val} set and 20k in the \textit{test-dev} set. We report analysis and ablation studies on the \textit{val} set and compare with other methods on the \textit{test-dev} set.

\subsection{Implementation Details}
\label{subsec:details}

We use ResNet-50~\cite{he2016deep} with FPN~\cite{lin2017feature} as backbone for all experiments if not specifically pointed out. We initialize the backbone with weights pre-trained on ImageNet~\cite{deng2009imagenet}. Following common practice, all models are trained for $1\times$ schedule named in \cite{he2019rethinking}, i.e., 90k iterations with an initial learning rate of 0.01, which is then divided by 10 at 60k and 80k iterations, with the weight decay of 0.0001 and the momentum of 0.9. Random horizontal flipping is used in data augmentation. For all ablations, we use an image scale of 800 pixels for training and testing, unless otherwise specified. We set $\tau=1/3$ in Eq.~\ref{eq:conf_weight_func}, and $\lambda=5.0$ in $\mathcal{P}(loc)$. Focal Loss with $\alpha=0.25$ and $\gamma=2.0$ is applied for negative classification. NMS with IoU threshold 0.6 is applied to merge the results.

\subsection{Ablation Studies}

\paragraph{Baseline.} None of existing label assignment strategies can be used as baseline of our AutoAssign, because we only rely on the \textit{center prior}, and do not require any other human knowledge like anchors, IoU thresholds, and top-$k$, which is indispensable for many other detectors. As a result, we build AutoAssign from a very simple and clean start point in Table~\ref{tbl:exp_locw_featw}. The 17.7 mAP baseline can be seen as removing $w^+$ and $w^-$ from Eq.~\ref{eq:loss_func}. Other detectors, like RetinaNet, can also be implemented by adding modules to this simple baseline.

\paragraph{Overall weighting mechanism.}


\setlength{\tabcolsep}{4pt}
\begin{table}[htbp]
\begin{center}
\begin{tabular}{c|c|c|cc|ccc}
\toprule
Center & Conf & $AP$ & $AP_{50}$ & $AP_{75}$ & $AP_S$ & $AP_M$ & $AP_L$ \\
\midrule
  & & 17.7 & 30.9 & 18.1 & 15.7 & 24.2 & 23.3 \\
  & \checkmark & 21.5 & 35.8 & 22.6 & 16.6 & 28.9 & 36.0 \\
\checkmark & &  37.7 & 57.4 & 40.6 & 20.3 & 41.4 & 52.0 \\
\checkmark & \checkmark & \bf40.5 & \bf59.8 & \bf43.9 & \bf23.1 & \bf44.7 & \bf52.9 \\
\bottomrule
\end{tabular}
\end{center}
\caption{Effectiveness of Center Weighting and Confidence Weighting. ``Center'' means center weighting, and ``Conf'' indicates confidence weighting. 
}
\label{tbl:exp_locw_featw}
\end{table}

To demonstrate the effectiveness of the two weighting modules, we construct the positives weights $w_i^+$ using only Center Weighting or Confidence Weighting separately in Table~\ref{tbl:exp_locw_featw}, while keeping the negatives weighting unchanged. Center Weighting brings relatively significant performance gain, suggesting that the prior distribution is critical for guiding the training. Besides, confidence weighting further improves the accuracy as it dynamically changes the strategy for each object in both spatial and scale dimensions according to object appearances. More design choices of the two modules can be found in Supplementary Materials.


\label{subsec:ablation}

\begin{figure*}[htbp]
	\centering
	\includegraphics[width=0.9\linewidth]{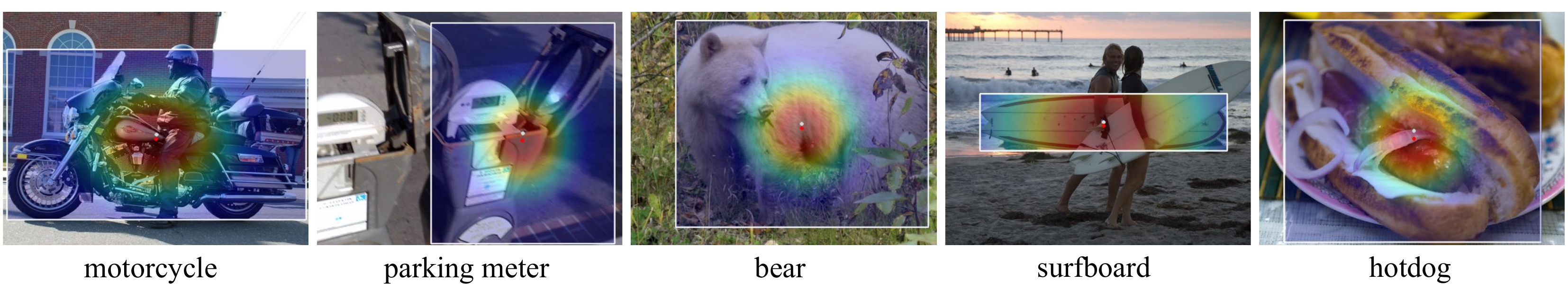}
	
\caption{Visualization of learned center weighting weights of different categories. All of the objects are visualized on the same scale. Center weighting results of motorcycle and surfboard show the prior distribution become ellipses (controlled by $\sigma$) to accommodate the shape characteristics of these categories.  Center offsets (controlled by $\mu$) are actually larger than 10 pixels in the raw image, which means it could shift one or more grids on output feature maps. Images are best viewed in color.}
\label{fig:da_lw_center}
\end{figure*}

To better understand how spatial and scale assignment is solved through the weighting mechanism, we visualize the positive weight maps separately in each FPN stage from a well-trained detector. From Fig.~\ref{fig:da_weighting}, the Center Weighting is applied to all FPN stages to achieve a coarse weighting based on the category-specific \textit{center prior}. Then the Confidence Weighting generates weights according to object appearances. The two modules perform spatial and scale assignments of each instance jointly.

\paragraph{Center Weighting.}

To analyze the design of the center weighting, we compare different prior distributions in Table~\ref{tbl:fix_and_dyn_locw}. We denote the Gaussian-shape function $G$ without learnable parameters as ``fixed'', while ``shared'' means all categories share one group of learnable $\vec{\mu}$ and $\vec{\sigma}$. Compared to the fixed prior, ``shared'' prior slightly drops the AP by 0.1\%, while our category-wise prior increases the AP by 0.2\% on MS COCO. As MS COCO contains 80 categories with a huge amount of data, its object distribution generally falls into a normal distribution. Thus the total improvement of category-wise prior is not significant. But when we look at some classes with unique distributions, \emph{e.g.}, surfboard, and hotdog, the improvements are notable.

\setlength{\tabcolsep}{4pt}
\begin{table}[htbp]
\begin{center}
\begin{tabular}{l|c|ccccc}
\toprule
Center & $AP$ & moto & prk-mtr & bear & surfboard & hotdog \\
\midrule
none & 21.5 & 15.2 & 14.9 & 66.3 & 7.9 & 11.5 \\
fixed & 40.3 & 42.2 & 41.9 & 71.9 & 32.4 & 33.5 \\
shared & 40.2 & 41.8 & 40.7 & 69.2 & 33.1  & 32.7 \\
\bf{category}  & \textbf{40.5} & \bf42.9 & \bf43.3 & \bf73.6 & \bf34.8 & \bf35.8 \\
\bottomrule
\end{tabular}
\end{center}
\caption{Results of different center weighting choices over the whole categories of MS COCO and the subset. ``moto'' means motorcycle, ``prk-mtr'' means parking meter.}
\label{tbl:fix_and_dyn_locw}
\end{table}

This can also be evidenced by the visualization of the learned priors for each category in Fig.~\ref{fig:da_lw_center}. We mark white points as the center of bounding boxes and red points as the center of learned priors. We can see that in the categories of parking meter and hotdog, the learned centers $\vec{\mu}$ shift down as these categories tend to have more essential clues at the bottom half. Moreover, the category-specific $\vec{\sigma}$ is also changed for each category. For the categories of motorcycle and surfboard, the prior becomes ellipses to accommodate the shape characteristics of these categories.

%

\paragraph{Confidence Weighting}

\begin{table}[htbp]
\begin{center}
\begin{tabular}{l|c|cc|ccc}
\toprule
Confidence & $AP$ & $AP_{50}$ & $AP_{75}$ & $AP_S$ & $AP_M$ & $AP_L$\\
\midrule
$\mathcal{P}(cls)$-only & 38.7 & \bf59.9 & 41.6 & 22.9 & 42.0 & 49.5 \\
$\mathcal{P}(loc)$-only  & 39.7 & 58.4 & 43.1 & 22.4 & 43.6 & 51.6  \\
\midrule
no-obj & 39.4 & 58.7 & 42.5 & 22.4 & 43.5 & 50.7 \\
explicit-obj & 39.5 & 58.8 & 42.3 & 21.6 & 43.4 & 52.2 \\
\midrule
AutoAssign & \textbf{40.5} & 59.8 & \textbf{43.9} & \textbf{23.1} & \textbf{44.7} & \textbf{52.9} \\
\bottomrule
\end{tabular}
\end{center}
\caption{Comparison of different choices for confidence weighting. $\mathcal{P}(cls)$-only means only use $\mathcal{P}(cls)$ for confidence weighting. ``no-obj'' means do not use \textit{ImpObj} for $\mathcal{P}(cls)$. ``explicit-obj'' means give the object-ness branch individual supervision, rather than sharing with classification.}
\label{table:exp_obj}
\end{table}


\begin{table*}[htbp]
\begin{center}
\begin{tabular}{l|c|c|cc|ccc}
\toprule
Method & Iteration & $AP$ & $AP_{50}$ & $AP_{75}$ & $AP_S$ & $AP_M$ & $AP_L$\\
\midrule
\multicolumn{8}{l}{\textbf{ResNet-101}}\\
RetinaNet~\cite{lin2017focal} & 135k & 39.1 & 59.1 & 42.3 & 21.8 & 42.7 & 50.2\\
FCOS~\cite{lin2017focal} & 180k & 41.5 & 60.7 & 45.0 & 24.4 & 44.8 & 51.6\\
FreeAnchor~\cite{zhang2019freeanchor} & 180k & 43.1 & 62.2 & 46.4 & 24.5 & 46.1 & 54.8\\
SAPD~\cite{zhu2019soft} & 180k & 43.5 & 63.6 & 46.5 & 24.9 & 46.8 & 54.6\\
ATSS~\cite{zhang2019bridging} & 180k & 43.6 & 62.1 & 47.4 & \textbf{26.1} & 47.0 & 53.6\\
AutoAssign (Ours) & 180k & \textbf{44.5} & \textbf{64.3} & \textbf{48.4} & 25.9 & \textbf{47.4} & \textbf{55.0}\\
\midrule
\multicolumn{8}{l}{\textbf{ResNeXt-64x4d-101}}\\
FCOS*~\cite{tian2019fcos} & 180k & 44.7 & 64.1 & 48.4 & 27.6 & 47.5 & 55.6\\
FreeAnchor~\cite{zhang2019freeanchor} & 180k & 44.9 & 64.3 & 48.5 & 26.8 & 48.3 & 55.9\\
SAPD~\cite{zhu2019soft} & 180k & 45.4 & 65.6 & 48.9 & 27.3 & 48.7 & \textbf{56.8}\\
ATSS~\cite{zhang2019bridging} & 180k & 45.6 & 64.6 & 49.7 & \textbf{28.5} & 48.9 & 55.6\\
AutoAssign (Ours) & 180k & \textbf{46.5} & \textbf{66.5} & \textbf{50.7} & 28.3 & \textbf{49.7} & 56.6\\
\midrule
\multicolumn{8}{l}{\textbf{ResNeXt-64x4d-101-DCN}}\\
SAPD~\cite{zhu2019soft} & 180k & 47.4 & 67.4 & 51.1 & 28.1 & 50.3 & 61.5\\
ATSS~\cite{zhang2019bridging} & 180k & 47.7 & 66.5 & 51.9 & 29.7 & 50.8 & 59.4\\
AutoAssign (Ours) & 180k & 48.3 & 67.4 & 52.7 & 29.2 & 51.0 & 60.3\\ 
AutoAssign (Ours)$\dagger$ & 180k & 49.5 & 68.7 & 54.0 & 29.9 & 52.6 & 62.0\\ 
AutoAssign (Ours)$\dagger\ddagger$ & 180k & \bf 52.1 & \bf 69.6 & \bf 58.0 & \bf 33.9 & \bf 54.0 & \bf 64.0\\ 
\bottomrule
\end{tabular}
\end{center}
\caption{Performance comparison with state-of-the-art one-stage detectors on MS COCO 2017 \textit{test-dev} set. All results listed adopt multi-scale training. $\dagger$ indicates multi-scale training with wider range $[480,960]$ used in~\cite{zhang2019freeanchor}. $\ddagger$ indicates multi-scale testing. * indicates improved versions.}
\label{table:exp-coco}
\end{table*}

\setlength{\tabcolsep}{4pt}
\begin{table*}[htbp]
\begin{center}
\begin{tabular}{l|c|cc|c|cc|c|cc}
\toprule
\multirow{2}{*}{Method} & \multicolumn{3}{c|}{PASCAL VOC} & \multicolumn{3}{c|}{Objects365} & \multicolumn{3}{c}{WiderFace} \\ \cline{2-10}
 & $AP$ & $AP_{50}$ & $AP_{75}$ & $AP$ & $AP_{50}$ & $AP_{75}$ & $AP$ & $AP_{50}$ & $AP_{75}$\\
\midrule
RetinaNet~\cite{lin2017focal} & 55.4 & 81.0 & 60.1 & 18.4 & 28.4 & 19.6 & 46.7 & 83.7 & 47.1\\
FCOS*~\cite{tian2019fcos} & 55.4 & 80.5 & 61.1 & 20.3 & 29.9 & 21.9 & 48.1 & 87.1 & 48.4\\
FreeAnchor~\cite{zhang2019freeanchor} & 56.8 & 81.1 & 62.1 & 21.4 & 31.5 & 22.8 & 46.3 & 81.6 & 47.5\\
ATSS~\cite{zhang2019bridging} & 56.6 & 80.7 & 62.6 & 20.7 & 30.0 & 22.4 & 48.9 & 87.1 & 49.7\\
AutoAssign (Ours) & \textbf{57.9} & \textbf{81.6} & \textbf{64.1} & \textbf{21.6} & \textbf{31.7} & \textbf{23.2} & \textbf{49.5} & \textbf{88.2} & \textbf{49.9}\\
\bottomrule
\end{tabular}
\end{center}
\caption{Performance comparison with typical detectors on PASCAL VOC, Objects365 and WiderFace. * indicates improved versions.}
\label{table:exp-generalization}
\end{table*}

%

We evaluate the effectiveness of classification confidence $\mathcal{P}(cls)$, localization confidence $\mathcal{P}(loc)$, and \textit{ImpObj} separately in Table~\ref{table:exp_obj}. In the first two rows, we respectively use classification confidence $\mathcal{P}(cls)$ and localization confidence $\mathcal{P}(loc)$ alone in the confidence weighting. The combination of the two confidences (AutoAssign) achieves higher performance, indicating that the joint confidence indicator is the preferable choice when evaluating a location's quality.

In the next two rows, we evaluate the contribution of \textit{ImpObj}. ``explicit-obj'' means that we explicitly supervise the \textit{objectness} branch with consistent labels (\emph{i.e.}, 1 for foregrounds and 0 for backgrounds) for all the locations inside the boxes. We find that simply using hard labels for the \textit{objectness} has no help to performance, while our \textit{ImpObj} can significantly boost the performance by $\sim$ 1\% AP. Moreover, the performance of objects at all sizes can obtain obvious performance gains. We think the contribution of \textit{ImpObj} comes from its effect on both filtering out the noise candidates and achieving better separation from the background. Visualizations can be found in Supplementary Materials.




\subsection{Comparison with State-of-the-art}
\label{subsec:sota}

We compare AutoAssign with other detectors on MS COCO \textit{test-dev} set. We adopt $2\times$ schedule following the previous works~\cite{tian2019fcos,zhang2019freeanchor,zhang2019bridging}. Results are shown in Table~\ref{table:exp-coco}. Under the same training setting, AutoAssign can consistently outperform other counterparts. For example, AutoAssign with ResNet-101 backbone achieves 44.5\% AP, and our best model achieves 52.1\% AP, which outperforms all existing one-stage detectors.

\subsection{Generalization}

Another benefit of using little human knowledge it that huge effort on hyper-parameters tuning when transfer to other datasets. To demonstrate the generalization ability, we evaluate AutoAssign and several other detectors on different data distributions, including general object detection (PASCAL VOC~\cite{pascal-voc-2007,pascal-voc-2012}, Objects365~\cite{shao2019objects365}) and face detection (WiderFace~\cite{yang2016wider}). In these experiments, we keep all the hyper-parameters unchanged and only adjust the training settings following the common paradigm of each dataset.

Results are shown in Table~\ref{table:exp-generalization}. We find that the performance of other methods with fixed or partly fixed assigning strategies are unstable on different datasets. Although they may achieve excellent performance on certain datasets, their accuracies on the other dataset may be worse. This proves that the label assignment strategy of these methods has low robustness, thus needs to be adjusted cautiously. In contrast, AutoAssign can automatically adapt to different data distributions and achieve superior performance without any adjustment.

\section{Conclusions}
In this paper, we propose a differentiable label assignment strategy named AutoAssign. It tackles label assignment in a fully data-driven manner by automatically determine the positives/negatives in both spatial and scale dimensions. 
It achieves consistent improvement to all the existing sampling strategies by $\sim$1\% AP with various backbones on MS COCO. Besides, extensive experiments on other datasets demonstrate that AutoAssign can conveniently transfer to other datasets and tasks without additional modification.

This new label assignment strategy also comes with new challenges. For example, the current weighting mechanism is not simple enough and can be further simplified, which will be left for future work.

{\small
\bibliographystyle{ieee_fullname}
\bibliography{egbib}

\begin{thebibliography}{10}\itemsep=-1pt

\bibitem{deng2009imagenet}
Jia Deng, Wei Dong, Richard Socher, Li-Jia Li, Kai Li, and Li Fei-Fei.
\newblock Imagenet: A large-scale hierarchical image database.
\newblock In {\em The IEEE Conference on Computer Vision and Pattern
  Recognition}, 2009.

\bibitem{pascal-voc-2007}
M. Everingham, L. Van~Gool, C.~K.~I. Williams, J. Winn, and A. Zisserman.
\newblock The {PASCAL} {V}isual {O}bject {C}lasses {C}hallenge 2007 {(VOC2007)}
  {R}esults.
\newblock
  http://www.pascal-network.org/challenges/VOC/voc2007/workshop/index.html.

\bibitem{pascal-voc-2012}
M. Everingham, L. Van~Gool, C.~K.~I. Williams, J. Winn, and A. Zisserman.
\newblock The {PASCAL} {V}isual {O}bject {C}lasses {C}hallenge 2012 {(VOC2012)}
  {R}esults.
\newblock
  http://www.pascal-network.org/challenges/VOC/voc2012/workshop/index.html.

\bibitem{he2019rethinking}
Kaiming He, Ross Girshick, and Piotr Doll{\'a}r.
\newblock Rethinking imagenet pre-training.
\newblock In {\em The IEEE International Conference on Computer Vision}, 2019.

\bibitem{he2016deep}
Kaiming He, Xiangyu Zhang, Shaoqing Ren, and Jian Sun.
\newblock Deep residual learning for image recognition.
\newblock In {\em The IEEE Conference on Computer Vision and Pattern
  Recognition}, 2016.

\bibitem{huang2015densebox}
Lichao Huang, Yi Yang, Yafeng Deng, and Yinan Yu.
\newblock Densebox: Unifying landmark localization with end to end object
  detection.
\newblock {\em arXiv preprint arXiv:1509.04874}, 2015.

\bibitem{kim2020probabilistic}
Kang Kim and Hee~Seok Lee.
\newblock Probabilistic anchor assignment with iou prediction for object
  detection.
\newblock {\em arXiv preprint arXiv:2007.08103}, 2020.

\bibitem{noisyanchor}
Hengduo Li, Zuxuan Wu, Chen Zhu, Caiming Xiong, Richard Socher, and Larry~S
  Davis.
\newblock Learning from noisy anchors for one-stage object detection.
\newblock {\em arXiv preprint arXiv:1912.05086}, 2019.

\bibitem{lin2017feature}
Tsung-Yi Lin, Piotr Doll{\'a}r, Ross Girshick, Kaiming He, Bharath Hariharan,
  and Serge Belongie.
\newblock Feature pyramid networks for object detection.
\newblock In {\em The IEEE Conference on Computer Vision and Pattern
  Recognition}, 2017.

\bibitem{lin2017focal}
Tsung-Yi Lin, Priya Goyal, Ross Girshick, Kaiming He, and Piotr Doll{\'a}r.
\newblock Focal loss for dense object detection.
\newblock In {\em The IEEE International Conference on Computer Vision}, 2017.

\bibitem{lin2014microsoft}
Tsung-Yi Lin, Michael Maire, Serge Belongie, James Hays, Pietro Perona, Deva
  Ramanan, Piotr Doll{\'a}r, and C~Lawrence Zitnick.
\newblock Microsoft coco: Common objects in context.
\newblock In {\em The European Conference on Computer Vision}, 2014.

\bibitem{liu2016ssd}
Wei Liu, Dragomir Anguelov, Dumitru Erhan, Christian Szegedy, Scott Reed,
  Cheng-Yang Fu, and Alexander~C Berg.
\newblock Ssd: Single shot multibox detector.
\newblock In {\em The European Conference on Computer Vision}, 2016.

\bibitem{redmon2016you}
Joseph Redmon, Santosh Divvala, Ross Girshick, and Ali Farhadi.
\newblock You only look once: Unified, real-time object detection.
\newblock In {\em The IEEE Conference on Computer Vision and Pattern
  Recognition}, 2016.

\bibitem{redmon2017yolo9000}
Joseph Redmon and Ali Farhadi.
\newblock Yolo9000: Better, faster, stronger.
\newblock In {\em The IEEE Conference on Computer Vision and Pattern
  Recognition}, 2017.

\bibitem{redmon2018yolov3}
Joseph Redmon and Ali Farhadi.
\newblock Yolov3: An incremental improvement.
\newblock {\em arXiv preprint arXiv:1804.02767}, 2018.

\bibitem{ren2015faster}
Shaoqing Ren, Kaiming He, Ross Girshick, and Jian Sun.
\newblock Faster r-cnn: Towards real-time object detection with region proposal
  networks.
\newblock In {\em Advances in Neural Information Processing Systems}, 2015.

\bibitem{rezatofighi2019generalized}
Hamid Rezatofighi, Nathan Tsoi, JunYoung Gwak, Amir Sadeghian, Ian Reid, and
  Silvio Savarese.
\newblock Generalized intersection over union: A metric and a loss for bounding
  box regression.
\newblock In {\em The IEEE Conference on Computer Vision and Pattern
  Recognition}, 2019.

\bibitem{shao2019objects365}
Shuai Shao, Zeming Li, Tianyuan Zhang, Chao Peng, Gang Yu, Xiangyu Zhang, Jing
  Li, and Jian Sun.
\newblock Objects365: A large-scale, high-quality dataset for object detection.
\newblock In {\em The IEEE International Conference on Computer Vision}, 2019.

\bibitem{tian2019fcos}
Zhi Tian, Chunhua Shen, Hao Chen, and Tong He.
\newblock Fcos: Fully convolutional one-stage object detection.
\newblock In {\em The IEEE International Conference on Computer Vision}, 2019.

\bibitem{wang2019region}
Jiaqi Wang, Kai Chen, Shuo Yang, Chen~Change Loy, and Dahua Lin.
\newblock Region proposal by guided anchoring.
\newblock In {\em The IEEE Conference on Computer Vision and Pattern
  Recognition}, 2019.

\bibitem{yang2016wider}
Shuo Yang, Ping Luo, Chen-Change Loy, and Xiaoou Tang.
\newblock Wider face: A face detection benchmark.
\newblock In {\em The IEEE Conference on Computer Vision and Pattern
  Recognition}, 2016.

\bibitem{yang2018metaanchor}
Tong Yang, Xiangyu Zhang, Zeming Li, Wenqiang Zhang, and Jian Sun.
\newblock Metaanchor: Learning to detect objects with customized anchors.
\newblock In {\em Advances in Neural Information Processing Systems}, 2018.

\bibitem{yu2016unitbox}
Jiahui Yu, Yuning Jiang, Zhangyang Wang, Zhimin Cao, and Thomas Huang.
\newblock Unitbox: An advanced object detection network.
\newblock In {\em The ACM International Conference on Multimedia}, 2016.

\bibitem{zhang2019bridging}
Shifeng Zhang, Cheng Chi, Yongqiang Yao, Zhen Lei, and Stan~Z Li.
\newblock Bridging the gap between anchor-based and anchor-free detection via
  adaptive training sample selection.
\newblock {\em arXiv preprint arXiv:1912.02424}, 2019.

\bibitem{zhang2019freeanchor}
Xiaosong Zhang, Fang Wan, Chang Liu, Rongrong Ji, and Qixiang Ye.
\newblock Freeanchor: Learning to match anchors for visual object detection.
\newblock In {\em Advances in Neural Information Processing Systems}, 2019.

\bibitem{zhu2019soft}
Chenchen Zhu, Fangyi Chen, Zhiqiang Shen, and Marios Savvides.
\newblock Soft anchor-point object detection.
\newblock {\em arXiv preprint arXiv:1911.12448}, 2019.

\bibitem{zhu2019feature}
Chenchen Zhu, Yihui He, and Marios Savvides.
\newblock Feature selective anchor-free module for single-shot object
  detection.
\newblock In {\em The IEEE Conference on Computer Vision and Pattern
  Recognition}, 2019.

\end{thebibliography}
}

\newpage

\appendix

\section{Design Choices of AutoAssign}

\subsection{Baseline}

To better understand the 17.7 AP Baseline in Table~\ref{tbl:fix_and_dyn_locw}  of our AutoAssign, we demonstrate how other methods can be implemented by adding corresponding elements onto it. For example, RetinaNet can be implemented by adding 9 \textit{anchors} to each FPN stage(so there are 9$\times$5 anchors in total), and corresponding positive and negative IoU threshold. FCOS\_imprv can be implemented by adding center sampling (CS) radius, artificial FPN scale assignment rules and \textit{center-ness} branch. Here we demonstrate how to add modules one by one to get the final FCOS\_imprv in Table~\ref{table:vanilla_to_fcos}.

\setlength{\tabcolsep}{4pt}
\begin{table}[htbp]
\begin{center}
\begin{tabular}{l|c}
\toprule
Method & $AP$ \\
\midrule
Baseline  & 17.4 \\
Baseline + CS  & 31.6 \\
Baseline + scale assignment & 34.0 \\
Baseline + CS + scale assignment & 37.7 \\ 
Baseline + CS + scale assignment + \textit{center-ness}  & 38.8 \\
\bottomrule
\end{tabular}
\end{center}
\caption{Build FCOS\_imprv on top of our Baseline step by step. The last row corresponds to official FCOS\_imprv. CS denots ``center sampling''.}
\label{table:vanilla_to_fcos}
\end{table}

\subsection{Initialization of Center Weighting}

We fix the learnable parameters $\mu$ and $\sigma$ to analyze if the learnable mechanism works and is useful for Center Weighting. We firstly set standard Gaussian parameters and find it works well. Then we change them to quite different parameters and find the performance goes down, which means LR are sensitive to these parameters. After that we release $\mu$ or $\sigma$ respectively, and we find that the performance recovers gradually, which means the learnable mechanism is working. Finally we initialize $\mu$ or $\sigma$ by the aforementioned values but make them learnable, which shows the learnable mechanism is slightly superior and robust to initialization.

\setlength{\tabcolsep}{4pt}
\begin{table}[htbp]
\begin{center}
\begin{tabular}{c|c|cc}
\toprule
Hyper-parameters & $AP$ & $AP_{50}$ & $AP_{75}$ \\ 
\midrule
fix $\mu=0,\sigma=1$ & 40.3 & 59.7 & 43.6 \\ 
fix $\mu=1,\sigma=2$ & 39.9 & 58.9 & 43.2 \\ 
fix $\mu=0$ & 40.2 & 59.4 & 43.5 \\ 
fix $\mu=1$ & 40.2 & 59.4 & 43.5 \\ 
fix $\sigma=1$ & 40.3 & 59.4 & 43.6 \\ 
fix $\sigma=2$ & 40.1 & 59.2 & 43.2 \\ 
init $\mu=0,\sigma=1$ (AutoAssign) & \textbf{40.5} & \textbf{59.8} & \textbf{43.9} \\ 
init $\mu=1,\sigma=2$ & 40.4 & 59.6 & 43.6 \\ 
\bottomrule
\end{tabular}
\end{center}
\caption{The results of AutoAssign when changing $\mu$ and $\sigma$. Bold fonts indicate the best performance}
\label{table:exp-location-fix}
\end{table}


\begin{table}
\begin{center}
\begin{tabular}{l|c|ccc}
\toprule
method & $f(x)$ & $AP$ & $AP_{50}$ & $AP_{75}$ \\ 
\midrule
$\text{exp}(x/\tau)$ & $\tau=1$  & 39.8 & 59.1 & 42.9 \\ 
$\text{exp}(x/\tau)$ & $\tau=1/3$ & \textbf{40.5} & \textbf{59.8} & \textbf{43.9} \\ 
$\text{exp}(x/\tau)$ & $\tau=1/5$ & 40.0 & 59.4 & 43.5 \\ 
\bottomrule
\end{tabular}
\end{center}
\caption{Different $\tau$ in the Confidence Weighting functions in AutoAssign. Bold fonts indicate the best performance.}
\label{table:exp-appearance-aware}
\end{table}

\begin{table*}
\begin{center}
\begin{tabular}{l|c|c|cc|ccc}
\toprule
Method & \textit{ImpObj} & $AP$ & $AP_{50}$ & $AP_{75}$ & $AP_S$ & $AP_M$ & $AP_L$\\
\midrule
\multirow{2}{*}{RetinaNet~\cite{lin2017focal}} &  & 35.9 & 55.9 & 38.2 & 19.8 & 39.5 & 47.9\\
 & \checkmark & 36.1 & 56.3 & 38.8 & 20.0 & 39.9 & 48.2\\
\midrule
\multirow{2}{*}{FCOS\_imprv~\cite{tian2019fcos}} &  & 38.8 & 57.6 & 42.2 & 22.7 & 42.9 & 50.0\\
 & \checkmark & 39.0 & 58.2 & 42.3 & 22.8 & 42.8 & 50.6\\
\midrule
\multirow{2}{*}{FreeAnchor~\cite{zhang2019freeanchor}} &  & 38.4 & 57.1 & 41.2 & 21.2 & 41.6 & 51.5\\
 & \checkmark & 38.5 & 57.6 & 41.5 & 20.9 & 41.9 & \textbf{53.1}\\
\midrule
\multirow{2}{*}{ATSS~\cite{zhang2019bridging}} &  & 39.3 & 57.8 & 42.6 & 23.0 & 43.0 & 50.7\\
 & \checkmark & 39.7 & 58.4 & 42.9 & 23.0 & 43.6 & 51.6\\
\midrule
\multirow{2}{*}{AutoAssign} &  & 39.4 & 58.7 & 42.5 & 22.4 & 43.5 & 50.7\\
 & \checkmark & \textbf{40.5} & \textbf{59.8} & \textbf{43.9} & \textbf{23.1} & \textbf{44.7} & 52.9\\
\bottomrule
\end{tabular}
\end{center}
\caption{The performances of the application of \textit{ImpObj} on different methods. Bold fonts indicate the best performance}
\label{table:exp-obj}
\end{table*}

\subsection{$\tau$ in Confidence Weighting}

Here we present the choices of Confidence Weighting functions in Table.~\ref{table:exp-appearance-aware}. We can find that neither too flat or steep slopes cann achieve the best performance. We guess that this hyper-parameter is similar to the $\gamma$ in the Focal Loss.

\subsection{Implicit-Objectness.}

We propose \textit{Implicit-Objectness} to help us to classify foreground or background. As shown in Table~\ref{table:exp-obj}, \textit{ImpObj} can boost AP on all detectors, but more significant on AutoAssign. The gain of \textit{ImpObj} in AutoAssign is $\sim$ 1.0 AP, much more than other detectors, shows that there is better interaction between the dynamic weighting mechanism and \textit{ImpObj}. We believe that the manual designed label assignment might conflict with the prediction of \textit{ImpObj} in some cases. For an instance, the center sampling assumes the areas near the center have the highest possibilities to locate on an object, which is unlikely in some cases like a bounding box of a giraffe, but \textit{ImpObj} will possibly tell us the center areas are background. On the other hand, applying $ImpObj$ on a small set of pre-selected positive samples might cause the effectiveness of it cannot be fully explored.

\section{Visualization}

\begin{figure*}[htbp]
\centering
\includegraphics[width=\linewidth]{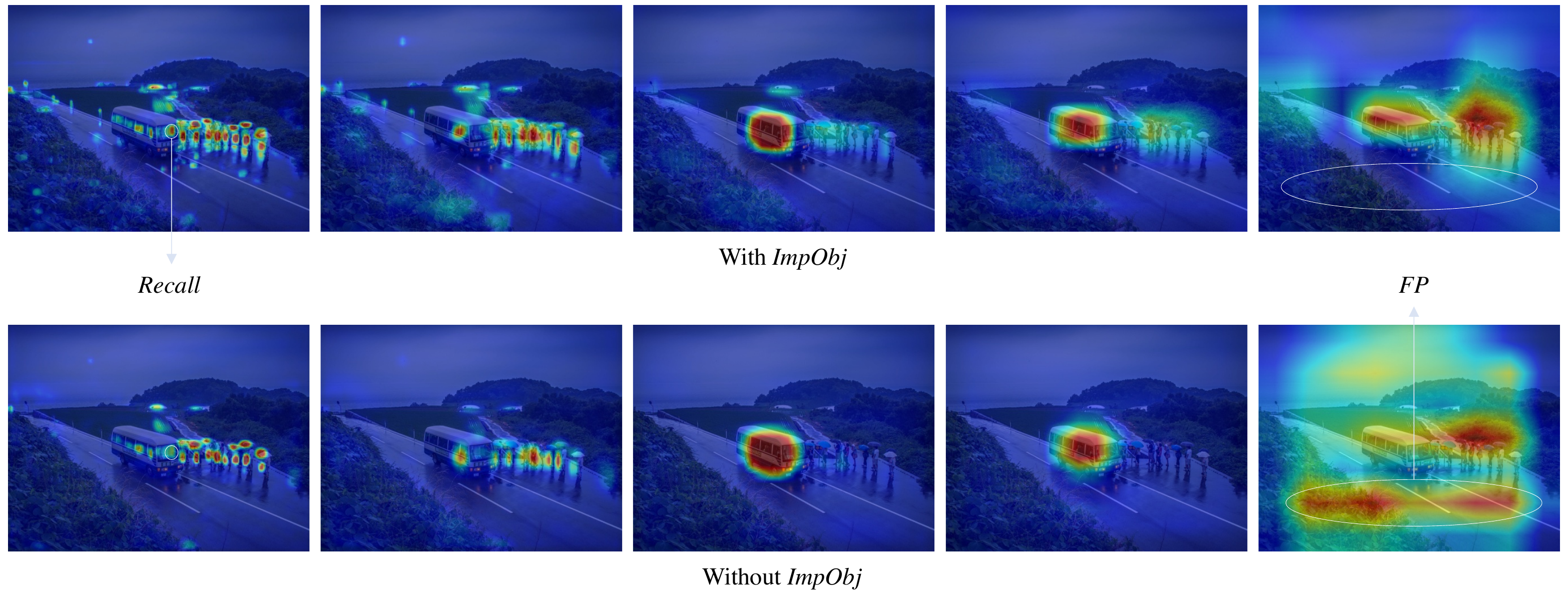}	
\caption{Confidence visualization of detectors trained with/without $ImpObj$. With the help of $ImpObj$, more objects can be detected, and false positives from background locations can be surpressed, thus recall and precision can be improved}
\label{fig:compare_cls_obj}
\end{figure*}

\begin{figure*}
\centering
\includegraphics[width=\linewidth]{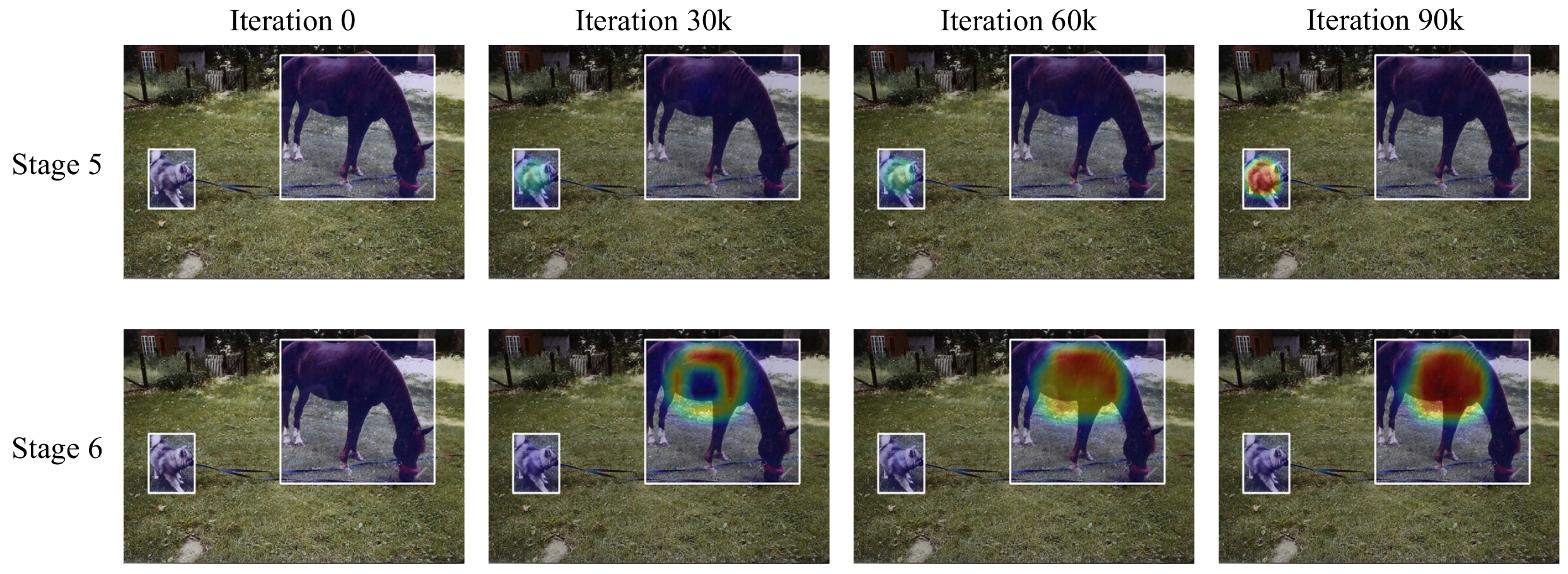}
\caption{Illustration of confidence weighting evolution from iteration 0 to 90k. We select the most activated stages for different objects}
\label{fig:da_weighting_trend}
\end{figure*}

To better understand the behavior of \textit{ImpObj}, we compare the final classification score used for NMS in Fig.~\ref{fig:compare_cls_obj}. From visualization and evaluation results, improvements come from both recall and precision. The proposed \textit{implicit-objectness} can filter out the noise and achieve better separation from background. Then we investigate the learning process of confidence weighting in Fig.~\ref{fig:da_weighting_trend}. We only visualize two representative FPN stages for simplicity. In the beginning, confidence weighting is weak for all objects because the probabilities among all locations are similarly low. With the training progress progressing, confidences become more salient and gradually converge to their appropriate FPN stages for objects of different sizes, demonstrating the effectiveness of our learnable strategy.

\end{document}